\documentclass[10pt,twocolumn,letterpaper]{article}

\usepackage{cvpr}              

\definecolor{cvprblue}{rgb}{0.21,0.49,0.74}
\usepackage[pagebackref,breaklinks,colorlinks,allcolors=cvprblue]{hyperref}
\usepackage{amsmath}

\usepackage[dvipsnames]{xcolor}

\def\confName{CVPR}
\def\confYear{2025}

\title{Flash-Split: 2D Reflection Removal with Flash Cues and \\ Latent Diffusion Separation}
\author{
    Tianfu Wang$^{}$\thanks{Equal Contribution.}  \quad Mingyang Xie$^{}$$^{*}$ \quad Haoming Cai$^{}$ \quad 
    Sachin Shah$^{}$  \quad Christopher A. Metzler$^{}$\\\\
    $^{}$University of Maryland 
}

\begin{document}
\maketitle

\begin{abstract}
{Transparent surfaces, such as glass, create complex reflections that obscure images and challenge downstream computer vision applications. {We introduce Flash-Split, a robust framework for separating transmitted and reflected light using a single (potentially misaligned) pair of flash/no-flash images. Our core idea is to perform latent-space reflection separation while leveraging the flash cues. Specifically, Flash-Split consists of two stages. Stage 1 separates apart the reflection latent and transmission latent via a dual-branch diffusion model conditioned on an encoded flash/no-flash latent pair, effectively mitigating the flash/no-flash misalignment issue. Stage 2 restores high-resolution, faithful details to the separated latents, via a cross-latent decoding process conditioned on the original images before separation.}
By validating Flash-Split on challenging real-world scenes, we demonstrate state-of-the-art reflection separation performance and significantly outperform the baseline methods.}
\end{abstract}

\begin{figure}[t]
  \centering
   \includegraphics[width=\columnwidth]{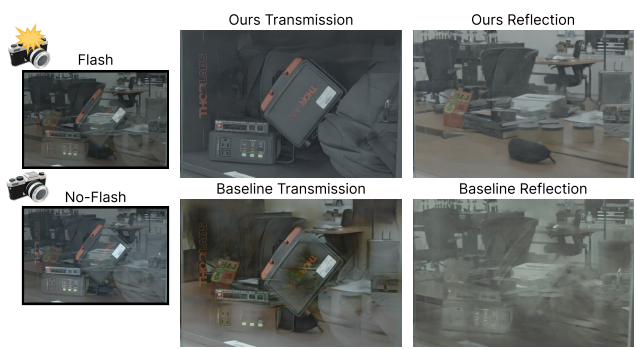}
   \caption{\textbf{Left}: We separated the transmitted and reflected scenes by capturing one image with camera flash and another with no flash, despite them being potentially misaligned due to hand shake. \textbf{Right}: Our proposed Flash-Split method archives a precise separation of the transmission and the reflection, performing much better than the baseline~\cite{lei2023tpami}. }
   
   \label{fig:1_teaser}
   \vspace{-20pt}
\end{figure}
\begin{figure}[t]
  \centering
   \includegraphics[width=\columnwidth]{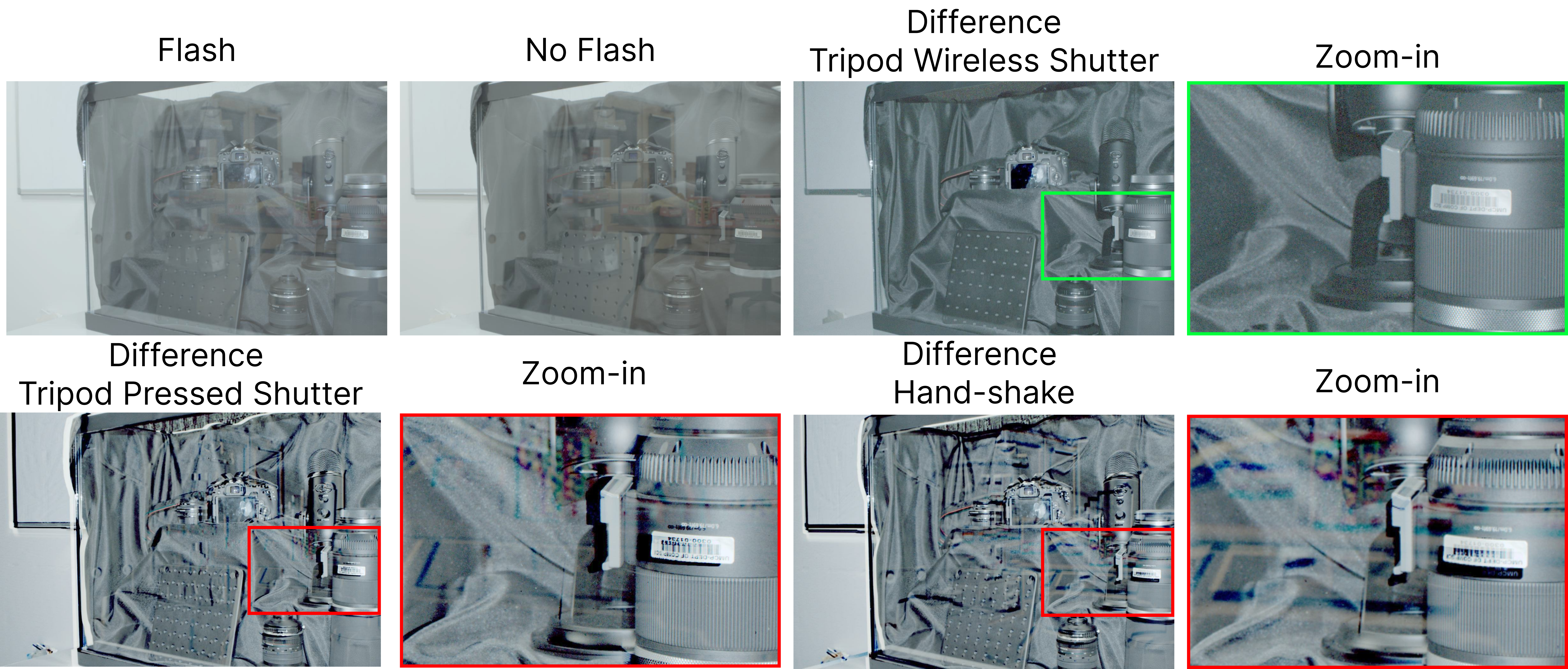}
   \vspace{-10pt}
   \caption{\textbf{Conventional Flash/No-Flash Methods Need Perfectly Paired Captures}. The camera flash increases the brightness of the transmitted scene without affecting that of the reflected scene. Therefore, the difference between this pair will be the transmitted scene free of reflection. \textbf{Top Right}: If we capture a perfectly aligned pair of flash/no-flash images using a tripod plus wireless shutter control, the difference is a perfect transmission image. \textbf{Bottom Left}: if we use a tripod but use a finger to press the shutter button, this slight motion will cause the two shots to be misaligned from each other, leading to noticeable artifacts in the difference image. \textbf{Bottom Right}: if we just do handheld photography, the difference image exhibits even stronger artifacts. \textbf{Takeaway}: this misalignment issue has been the key barrier to applying flash/no-flash photography, an accessible method with great potential,  to the task of reflection removal. In our work, we propose a robust approach to circumvent this key barrier.}
   \label{fig:2_motivation}
   \vspace{-10pt}
\end{figure}
\begin{figure}[t]
  \centering
   \includegraphics[width=0.9\columnwidth]{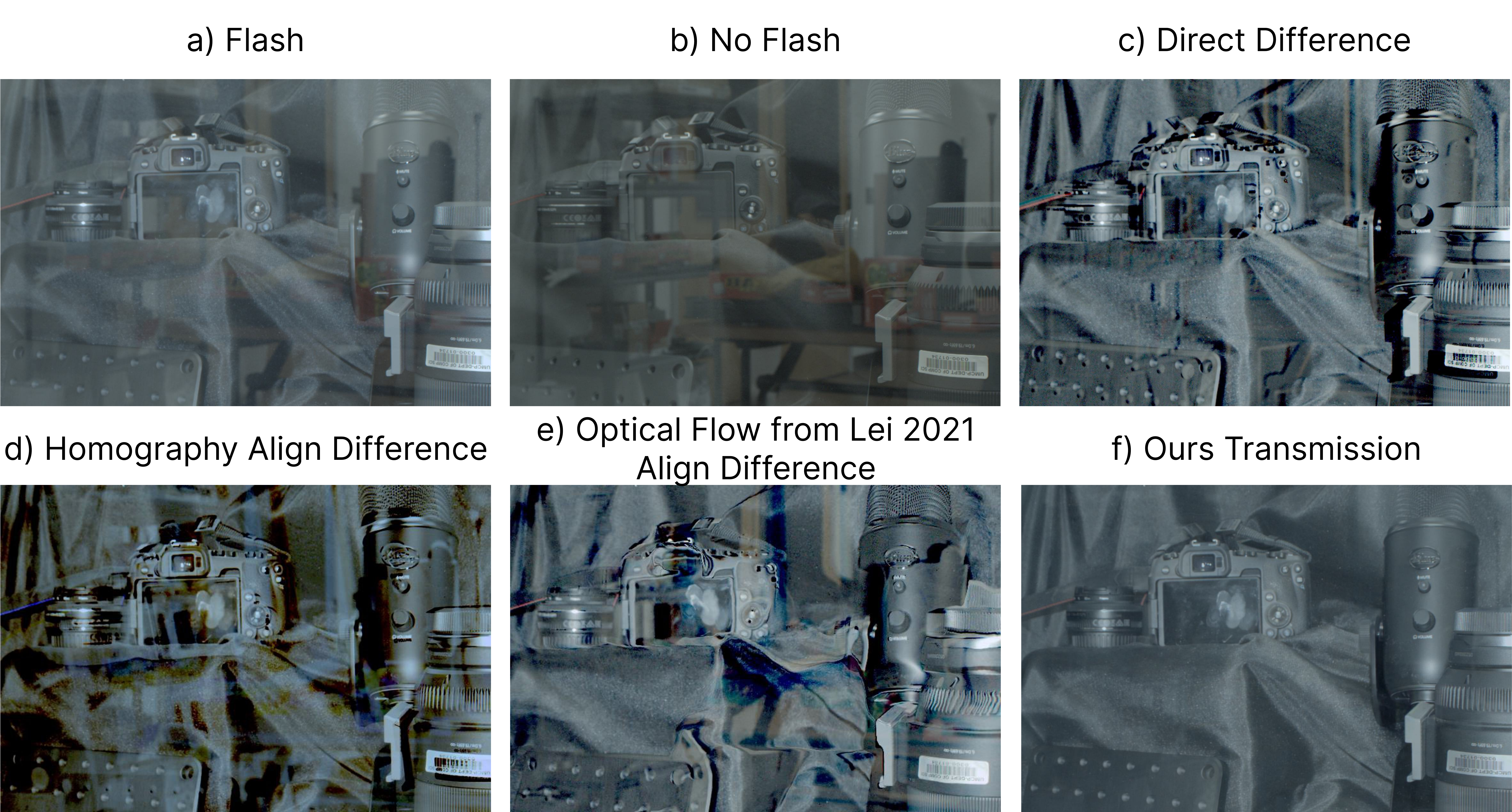}
   \vspace{-2pt}
   \caption{\textbf{Aligning Flash/No-Flash Images Is A Difficult Task for Image Registration Methods}. While the difference between a misaligned flash/no-flash image pair (\textcolor{red}{a},\textcolor{red}{b}) exhibits severe artifacts (\textcolor{red}{c}), aligning them is a non-trivial problem, since camera flash modifies the appearance of the transmitted component of one of the two images. Existing registration methods, like homography (\textcolor{red}{d}~\cite{fischler1981random}) or optical flow prediction (\textcolor{red}{e}~\cite{Sun2018PWC-Net}) used in~\citet{lei2023tpami}, fail to align this pair of images well -- their aligned flash/no-flash pair still suffer from severe artifacts. In contrast, our method (\textcolor{red}{f}) circumvents the misalignment issue by directly encoding the flash/no-flash pair into the latent space to perform recursive latent separation, eventually yielding a clean transmission scene.
   }
   \label{fig:3_alignment}
   \vspace{-10pt}
\end{figure}
\section{Introduction}
\label{sec:intro}

{Scenes with transparent surfaces, especially glass, frequently surround us and create specular reflections. In such scenes, what we perceive is a combination of transmitted and reflected light. This study aims to separate the transmitted and reflected 2D scenes.}

{Reflection removal and separation have garnered significant interest in the low-level vision community~\cite{levin2002learning, zhu2024revisiting, xie2024flash-splat, lei2020polarized, lei2021robust}. By removing the reflections in the scene, we can not only enhance visual quality but also boost the performance of downstream vision tasks such as depth estimation, robot navigation, object classification, and scene understanding.} 

Separating the reflection from the transmission is a challenging task due to its highly under-determined nature.
With both the transmission and reflection being unknown, it is  challenging to solve for each of them just based on their summed intensities.
To overcome this challenge, existing approaches have leveraged various prior assumptions.
Some approaches, for example, assume the reflection is out of focus~\cite{arvanitopoulos2017single,yang2019fast} or that the front and back sides of the glass cause significant double reflection~\cite{shih2015reflection}. However, these assumptions might not always hold in real-world scenarios. 

On the other hand, adding illumination control, e.g., using the built-in flash of a camera, is both accessible to everyday users and demonstrates significant potential in reducing the under-determined nature of reflection separation.
Specifically, the flash/no-flash technique~\cite{Agrawal2005flash_vanilla} performs reflection separation by capturing two images from the same viewpoint: one image with the camera flash on and another image with the camera flash off. While the camera flash boosts the intensity of the transmitted scene, it mostly leaves the reflected scene unchanged (more details in Section~\ref{subsec:method_intuition}). Consequently, by subtracting the image captured without flash from the image captured with flash, we can retrieve a transmission scene free of reflections~\cite{lei2021robust}.

The primary limitation of this approach is its reliance on precisely aligned flash/no-flash image captures, meaning the camera must remain stationary between shots. Even minimal movements, like pressing the shutter, can misalign the image pair, rendering this approach ineffective (\cref{fig:2_motivation}). To use this approach, users either need to hold their hands perfectly still during the two-shot capture, which is realistically infeasible for humans, or use a tripod plus remote shutter control. This requirement for paired captures poses a significant challenge for effective reflection separation in uncontrolled, real-world conditions.

To overcome this limitation, \citet{lei2023tpami} explored pre-align the flash image and the no-flash image via an optical flow module~\cite{Sun2018PWC-Net}. However, aligning the flash and no-flash images is much more difficult than the usual optical flow/homography task since the flash modifies the appearance of the objects in the transmitted scene. From our empirical experiments, we found that such pre-alignment is not robust when evaluating on diverse real-world flash/no-flash images (\cref{fig:3_alignment}). 

In our paper, we develop a novel approach for robust reflection separation, using a pair of misaligned flash/no-flash images. 
The key idea in this paper is to leverage flash cues to perform \textit{latent-space} reflection separation.
Our intuition behind it is that the condensed latent space makes it easier for our model to perform reflection separation under the flash guidance, while being more robust to the flash/no-flash misalignment issue.
Guided by this idea, we further decouple the reflection separation problem into two consecutive stages: (1) recursive latent separation and (2) cross-latent decoding.

In Stage 1 of our method, given a potentially misaligned {flash/no-flash image pair} as input, we first encode it into a {flash/no-flash latent pair} using a VAE encoder~\cite{Kingma2013VAE}; afterward, we recursively separate apart the latent representations for the reflected scene and the transmitted scene, using a dual-branch diffusion model conditioned on the flash/no-flash latent pair. 
By implicitly leveraging the flash cues at latent space, our dual-branch diffusion learns to effectively distinguish between the features from the transmitted scene and those from the reflected scene.

However, while we can effectively separate apart the transition and reflection in the latent space, naively decoding them to RGB space will tend to hallucinate content details (especially the high-frequency details) due to the inevitable under-determinedness of the decoding process, therefore reducing the faithfulness of the final separated images. To solve this, we introduce our Stage 2 \text{cross-latent decoding}, where we use the separated latent as guidance to extract the sharp image features from the unseparated input image. On a higher level, we are fusing the well-separated yet highly condensed information in our predicted latents with the unseparated yet highly detailed input image, to reconstruct a well-separated image that preserves fine and faithful details. By combining Stage 1 and Stage 2, our proposed method significantly outperforms baselines and has been validated on challenging real-world scenes.

Our contributions are:
\begin{itemize}
    \item We propose Flash-Split, a robust 2D reflection removal framework that combines flash/no-flash physical cues and latent space transmission/reflection separation.
    \item We develop a dual-branch diffusion framework for recursive latent separation, which can effectively handle misaligned flash/no-flash input images. 
    \item We use a cross-latent decoding module to restore faithful and high-frequency details from our separated latents.
    \item We demonstrate that our approach effectively separates reflection and transmission in real scenes, outperforming all baselines, including other flash/no-flash-based methods, in challenging cases with strong reflections.

\end{itemize}

\section{Related Works}
\label{sec:related}

Reflection removal has been a long-standing task in computational photography, with existing methods roughly categorized into three general categories: software-only, multi-view, and hardware-based. 

\noindent\textbf{Software-only Reflection removal.} The majority of software-only reflection removal works
take only one single image with mixed transmission and reflection components and attempt to separate the two.
Traditionally, this involves using prior statistical properties of the reflection~\cite{levin2002learning, levin2004separating, levin2007user, li2013exploiting}.
Recently, deep learning methods~\cite{,shih2015reflection,hariharan2015hypercolumns,wan2018region,yang2019fast,arvanitopoulos2017single,fan2017generic,yang2018seeing,zhang2018single,li2020single,wan2018crrn,wan2021face,wen2019single,wei2019single,kim2019single,zou2020deep,zheng2021single,dong2021location,hu2023single,li2023two,hu2021trash,wan2020reflection,Kee2024RawReflection, wang2024promptrr, zhu2024revisiting} have emerged where the reflection can be learned in a data-driven manner with promising results. Nevertheless, given the inherent ill-posed nature of reflection removal, software-only approaches are not robust to complex real world scenarios, especially in scenes with strong reflections.

\noindent\textbf{Multi-View Reflection Removal.} Multi-frame approaches~\cite{farid1999separating,gandelsman2019double,hong20232,hong2021panoramic,liu2020learning,alayrac2019visual,xue2015computational,guo2014robust,li2013exploiting,chugunov2024nsf, xie2024flash-splat, guo2022nerfren} aims to combine temporal and spatial cues for consistent reconstruction and separation. 
 Among them, unsupervised works such as NeRFRen~\cite{guo2022nerfren} use neural fields to model both the transmitted and reflected 3D scenes, leveraging cross-view consistency as cues for 3D reflection separation. 

\noindent \textbf{Hardware-Related Reflection Removal}. 
These methods introduce hardware elements to exploit optical cues of the transmission and reflection light transport.
Some studies employ polarization cues~\cite{lei2020polarized, nayar1997separation, kong2011high,kong2013physically, lyu2019reflection, li2020reflection}. They leverage the fact that the transmission is unpolarized while the reflection component varies when rotating the polarization filter. 
However, access to polarization cameras is limited to general camera/smartphone users.
Other works, including our paper, involve taking a pair of flash/no-flash images from the same view point~\cite{lei2021robust,Xia2020FlashNormal,Xia2020FnF}, the mechanism of which will be introduced in detail in the next section. 

\begin{figure*}[t]
  \centering
   \includegraphics[width=0.9\textwidth]{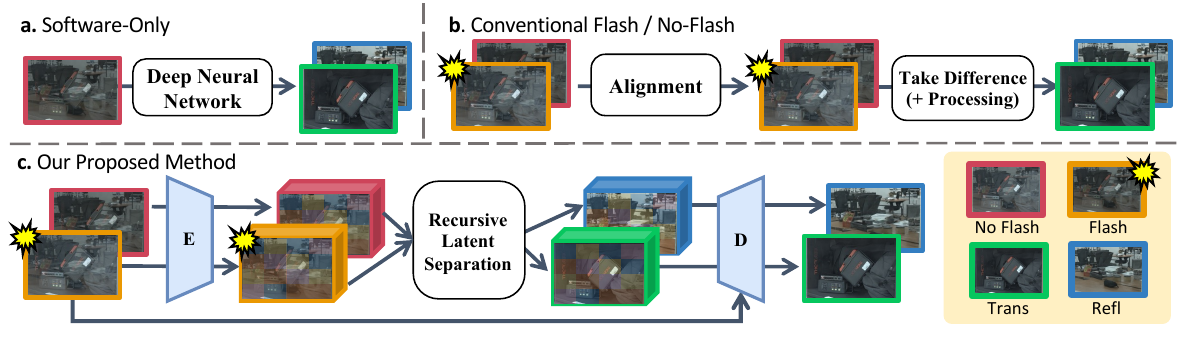}
   \vspace{-5pt}
   \caption{\textbf{Comparing Different 2D Reflection Removal Paradigms.}
   (\textcolor{red}{a}): \textbf{Software-only methods} pass a single composite image (with both transmission and reflection) to a deep neural net for reflection separation. (\textcolor{red}{b}): \textbf{Conventional flash/no-flash methods} take the difference of a flash/no-flash image pair to get the transmission image~\cite{Agrawal2005flash_vanilla}; optionally, one can also use a neural net~\cite{chang2020siamese} to predict the reflection image and further refine the transmission image quality (omitted in the figure for simplicity). In cases of misalignment (when not using a tripod), \citet{lei2023tpami} uses an optical flow module to pre-align the image pair. (\textcolor{red}{c}): \textbf{Our proposed method} encodes the flash/no-flash method down to the latent space: we first encode the flash/no-flash image pair into a flash/no-flash latent pair, then use its physical cue to separate the composite scene's latent into a transmission latent and a reflection latent, and finally decode them back to RGB image space to obtain the clean transmission image and reflection image. 
   }
   \vspace{-5pt}
    \label{fig:4_overview}
\end{figure*}
\begin{figure*}[t]
  \centering
   \includegraphics[width=0.9\textwidth]{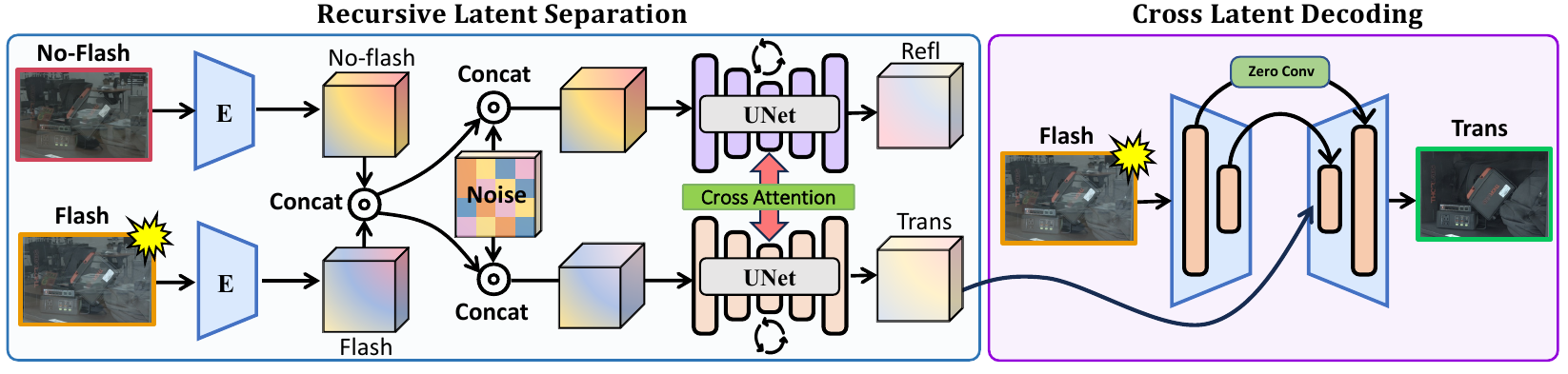}
   \vspace{-5pt}
   \caption{\textbf{Our Proposed Pipeline}  consists of a latent separation stage and a decoding stage. \textbf{Left:} We first encode the misaligned flash/no-flash image pair into a flash/no-flash latent pair. We then use a dual-branch attention UNet with cross-attention in-between to perform latent separation --- the goal is to predict a latent for the transmission scene and another latent for the reflection scene. Following recent development of latent diffusion models~\cite{rombach2021latentdiffusion, ke2023marigold}, at each inference step, we concatenate both the flash/no-flash latents with random Gaussian noise and let the dual-branch UNet denoise them. Eventually, the top and bottom branches predict a transmission and reflection latent, respectively. \textbf{Right:} We observe that the vanilla decoding process may lead to hallucination and blurriness~(Figure \ref{fig:12_decoder}). To fix this issue, we apply a cross-latent decoding process with a UNet~\cite{unet} architecture. But unlike a normal UNet, we do not feed the encoder's output into the decoder. Instead, we (\textcolor{red}{1}) feed the original unseparated image into the encoder and (\textcolor{red}{2}) feed our separated latent (from the first stage) directly into the decoder. The encoder passes information to the decoder only through the skip connection layers. This decoding process combines two complementary sources of information: the predicted latent from Stage 1, separated but missing high-frequency information, and the captured image, unseparated but contains high-frequency details, leading to a faithful reconstruction of the original transmission/reflection scenes. 
}

   \label{fig:5_pipeline}
\end{figure*}
\section{Proposed Method}
\label{sec:method}
\subsection{Flash/No-Flash Preliminaries}
\label{subsec:method_prelim}
An established technique~\cite{lei2021robust} among photographers to obtain a reflection-free image is to compare images taken with and without flash from the same viewpoint. 
Assume we have a composite scene consisting of a transmission-only scene $\mathbf{T}$, a reflection-only scene $\mathbf{R}$, and a transparent reflective surface, e.g., glass.
The image of this composite scene $\mathbf{I}$ can be formulated as
\begin{align}
    \mathbf{I} &=  \mathbf{T}  + \gamma\circ \mathbf{R}
    \label{eqn:noflash}
\end{align}
Now, assume we take a second image from the exact same viewpoint as the first image, only now turning on an additional illumination source co-located on the viewpoint, e.g., a camera flash. 
Assuming the illumination (camera flash) strength is uniformly distributed, it will increase the intensity across all pixels in the transmission scene in proportion to each pixel's reflectivity. 
Additionally, if the glass is not perpendicular to the camera viewing direction, the flash illumination will reflect away from the camera sensor after it hits the reflective surface, avoiding any flares on the glass in the captured image.

While the flash illumination will cause secondary reflections (e.g., light that hits the glass, bounces to the reflected scene, and then gets reflected elsewhere), the chances of secondary reflections getting back to the camera sensor are very low.
Therefore, it is reasonable to assume that there are little changes to the intensity of the reflection scene when we use flash illumination.
Under these conditions, we can now approximate of the image taken with flash $\mathbf{I}_{Flash}$ as
\begin{align}
    \mathbf{I}_{Flash}  &\approx  (1 + \theta)\mathbf{T} + \gamma \circ \mathbf{R},
\end{align}

Taking the difference between the flash image $\mathbf{I}_{Flash}$ and no-flash image $\mathbf{I}$, we shall obtain an image of the transmitted scene:
\begin{align}
    \mathbf{I}_{Flash} - \mathbf{I}  &\approx \theta \circ \mathbf{T}.
\end{align}

A visual example of how this approach works is shown in \cref{fig:2_motivation} top row. Note that this difference image will slightly differ from the transmitted scene when there is no glass, because the intensity of the difference image is dependent on the flash illumination strength.

\subsection{Our Core Idea}
\label{subsec:method_intuition}

While the flash/no-flash approach has great potential for reflection removal, it requires a \textit{perfectly aligned} pair of images, e.g., from the same camera viewpoint. Otherwise, the flash/no-flash difference will contain heavy artifacts. 
As shown in the bottom two rows of \cref{fig:2_motivation}, any motion during capture, including user hand shake or even pressing the shutter button, will cause this method to break. 

To overcome the misalignment issue, the most straightforward way is to align the image pair first before taking the difference, which has been explored by \citet{lei2023tpami}. As shown in \cref{fig:3_alignment}, the difference image after alignment by homography~\cite{fischler1981random} or optical flow~\cite{Sun2018PWC-Net} still suffers from noticeable artifacts, not only because the flash/no-flash method is very sensitive to alignment error, but also because the flash changes the appearance of the transmitted scene, making it harder for registration methods to work.

In our work, we take an alternative path to deal with the misalignment issue. Inspired by recent works on image latent features~\cite{rombach2021latentdiffusion, Kingma2013VAE, Phung2023attrefocus}, we take the flash/no-flash method down to the latent space. Similar to how conventional flash/no-flash methods~\cite{Agrawal2005flash_vanilla} take advantage of a flash/no-flash image pair, we create a flash/no-flash \textit{latent pair} via a variational autoencoder (VAE)~\cite{Kingma2013VAE}. Then, we perform latent-space reflection separation using this flash/no-flash latent pair, to obtain one latent for the transmission image, and another latent for the reflection image. In the end, we decode the separated latents back to RGB space to obtain the transmission/reflection images. Below we explain why we choose to do separation in latent space.

\subsubsection{Latent Separation Mitigates Misalignment}
When a vision encoder~\cite{Kingma2013VAE} encodes an image, it expands the feature dimension while reducing the spatial dimension. By focusing on the overall high-level features rather than precise pixel locations, the encoder effectively reduces the impact of spatial misalignment between the flash/no-flash image pair. Specifically, when performing feature extraction, the encoder also aggregates local pixel information, averaging out the difference in misaligned pixel location.

\vspace{-3pt}

\subsubsection{Reflection Separation Is Easier in Latent Space}
Our key intuition here is that training a model to separate the composite scene's latent into the transmitted scene's latent and the reflected scene's latent will be much easier than training a model to separate a composite image into a transmission image and a reflection image.

More specifically, the high-level representation of image latents allows a model to better focus on separating the main features in the reflected and transmitted scenes(such as the primary object structures). Training with reduced dimensionality also lets a model converge to a better local optimum and have better generalization ability. Once the separation is done in latent space, we then use our customized decoder to restore the fine details, reconstructing a sharp and well-separated image (Further discussed  in \cref{subsec:method_stage2}).

\subsubsection{Leveraging Flash Cues in The Latent Space}
Similar to how the flash/no-flash technique works in the RGB image space (\cref{subsec:method_prelim}), after we encode the flash/no-flash image pair into a latent pair, it can still provide important cues for reflection separation, despite being at a condensed latent space. This is because the physical cues from the flash/no-flash technique lie in a relatively low-frequency domain. 
Intuitively, suppose we have a flash/no-flash image pair with the difference being the transmission; if we downsample them to a smaller dimension, their difference image, despite being low-resolution, would still resemble the transmission scene, serving as a powerful cue.

An ablation is shown in \cref{fig:14_single_input}. While keeping the reflection separation happening in latent space, we compare two model variants: one takes flash/no-flash as input, and the other takes in a single image as input. It turns out the former significantly outperforms the latter, implying that the flash/no-flash cues can still be leveraged in latent space.

\begin{figure}[h]
  \centering
   \includegraphics[width=0.9\columnwidth]{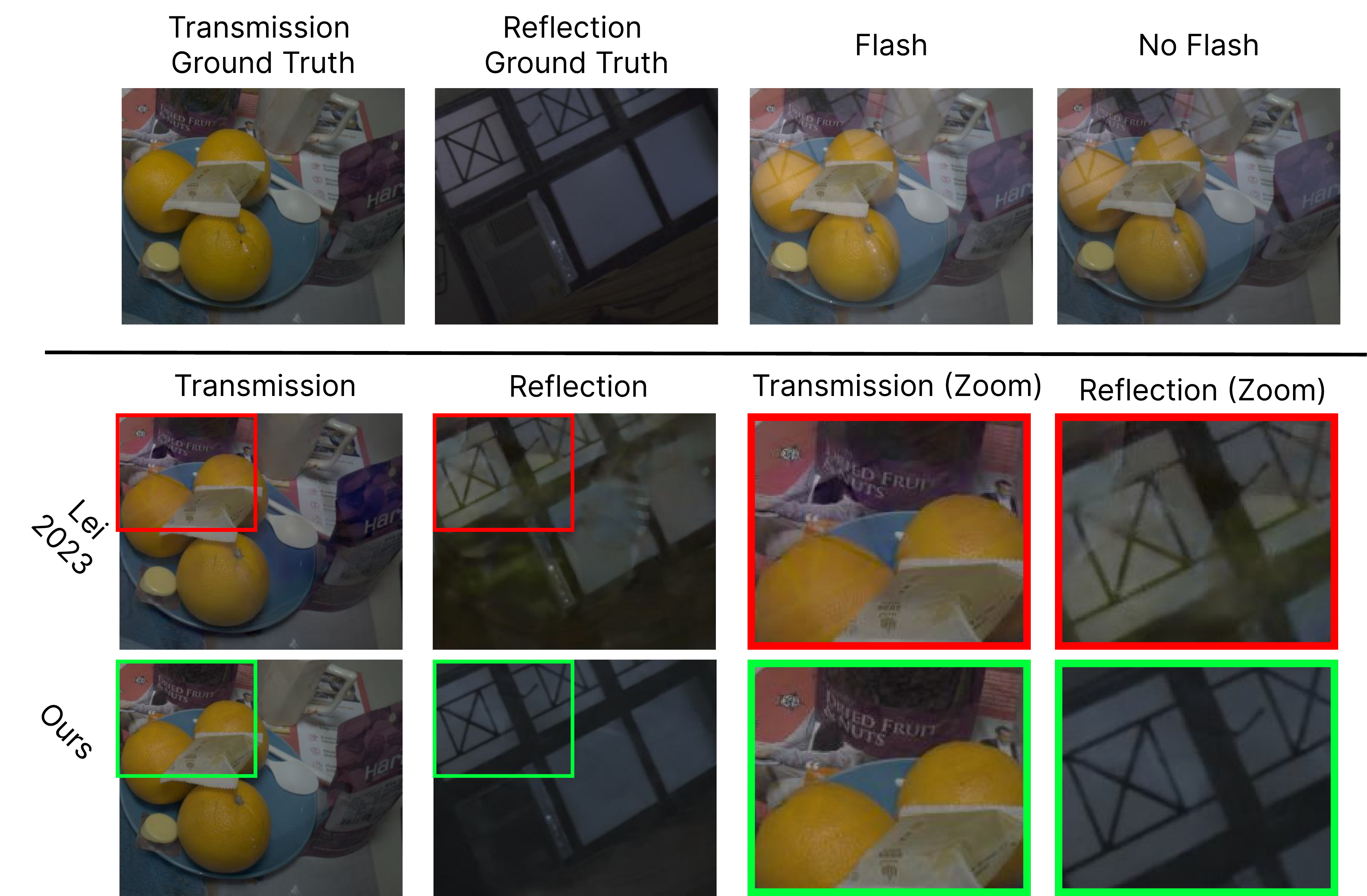}
   \vspace{-5pt}
   \caption{\textbf{Qualitative Comparison With Ground Truth Transmission.} We qualitatively compare our method with \citet{lei2023tpami} on the dataset introduced by themselves, which contains the ground truth transmission captured by removing the glass that causes the reflection. While our model is trained on exactly the same datasets as theirs, our model performs better reflection separation, due to our latent separation strategy. In addition, note that the reflection strength of scenes in this dataset is very weak, making it easy to separate. }
   \label{fig:6_sim}
 \vspace{-10pt}
\end{figure}

\begin{figure*}[h!tb]
  \centering
   \includegraphics[width=0.9\textwidth]{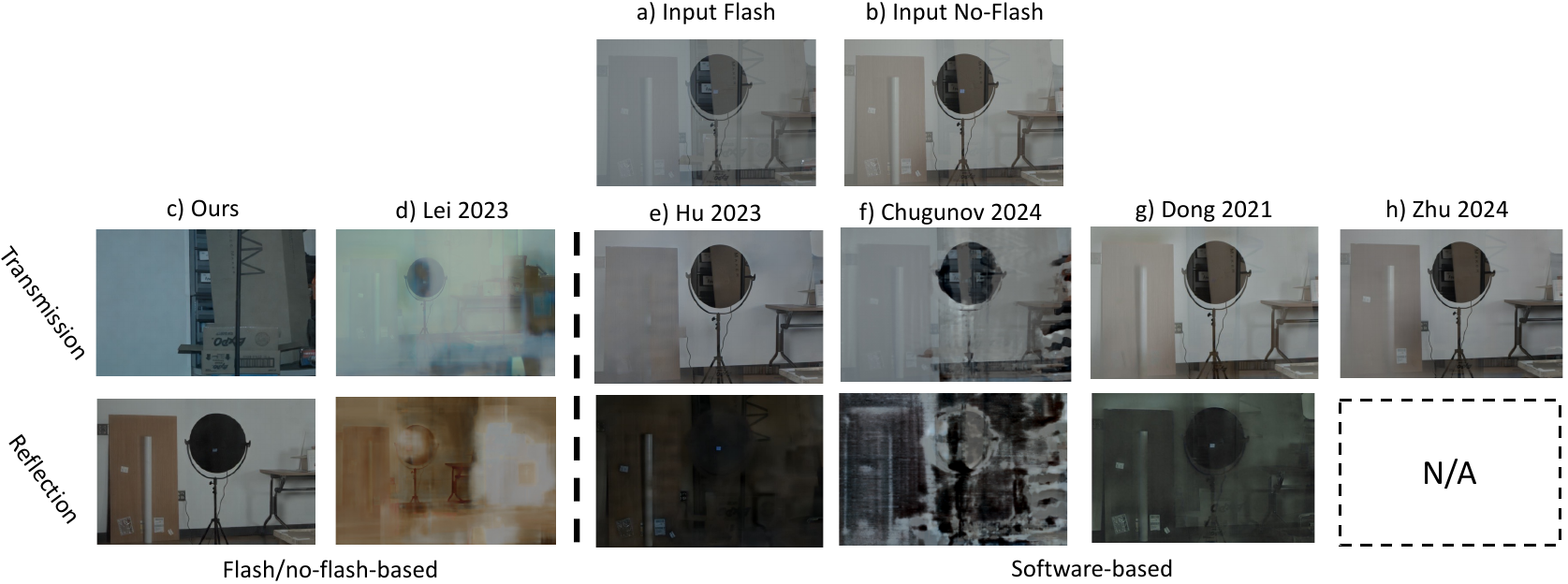}
      \vspace{-5pt}

   \caption{\textbf{Real Experiment: The Lab Scene.}  The transmission is some paper boxes; the reflection is a door and lamp. Using the top-row flash/no-flash image pair (misaligned due to motion between shots), our method overcomes the misalignment and achieves reflection separation not only better than software-only approaches (\textcolor{red}{e},\textcolor{red}{f},\textcolor{red}{g},\textcolor{red}{h})~\cite{hu2023single, chugunov2024nsf, dong2021location, zhu2024revisiting}, but also better than another flash/no-flash based method (\textcolor{red}{d})~\cite{lei2023tpami}, which is trained on exactly the same dataset as ours. Note that \citet{zhu2024revisiting} can only predict the transmission, not the reflection, thus the ``N/A''.
   }
   \vspace{-5pt}
   \label{fig:7_main_lab}
\end{figure*}
\begin{figure*}[ht]
  \centering
   \includegraphics[width=0.9\textwidth]{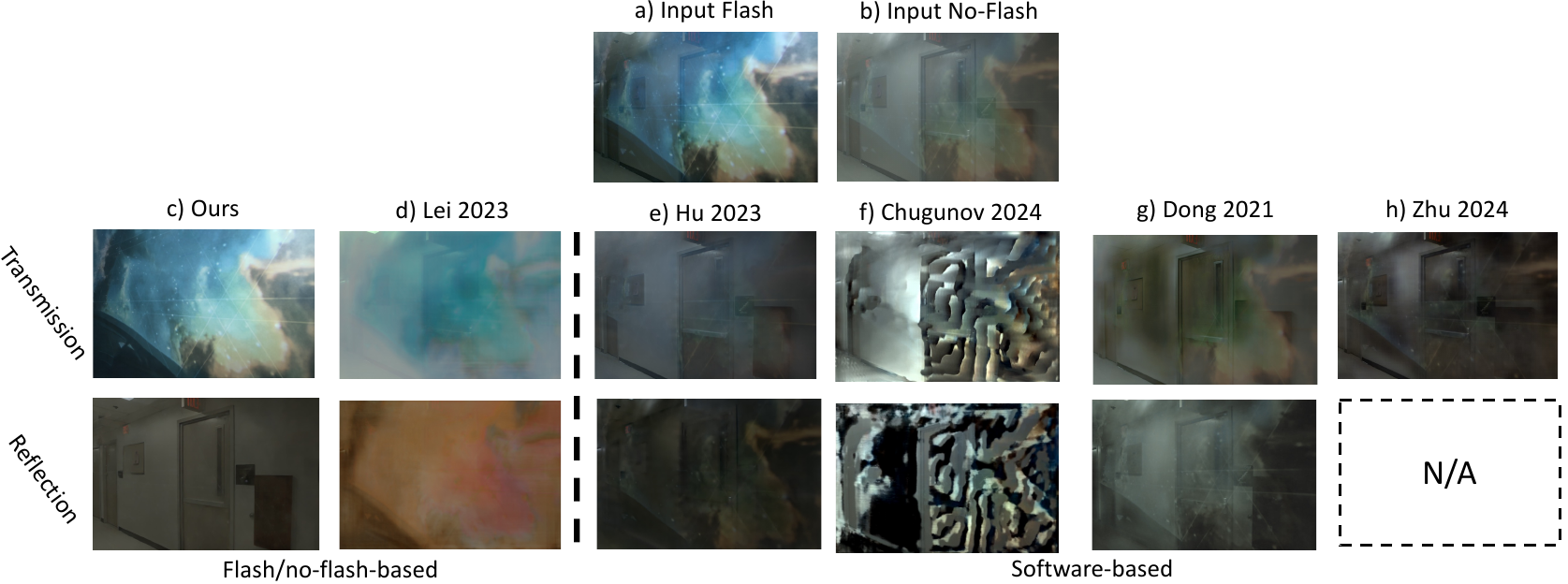}
   \vspace{-5pt}
   \caption{\textbf{Real Experiment: The Poster Scene.} The transmission is a poster; the reflection is a hallway.
   Our method achieves superior reflection separation than all the baselines (\textcolor{red}{d},\textcolor{red}{e},\textcolor{red}{f},\textcolor{red}{g},\textcolor{red}{h})~\cite{lei2021robust, hu2023single, chugunov2024nsf, dong2021location, zhu2024revisiting}}
   \label{fig:7_main_poster}
   \vspace{-10pt}
\end{figure*}
\subsection{Our Pipeline}
\label{subsec:method_pipeline}

Following our core idea described in the previous section, we decouple the reflection separation problem into two sub-problems: (\textcolor{red}{1}) after we first encode the composite (flash/no-flash) images into latents, how to separate them into a latent representation for the transmission image and the reflection image, respectively; and (\textcolor{red}{2}): how to restore high-frequency details to the separated latents while keeping the details faithful to the original scene.
To address the two issues, we propose a 2 stage framework consisting of recursive latent separation and cross-latent decoding, described in Sec.~\ref{subsec:method_stage1} and \ref{subsec:method_stage2}, respectively.
To better illustrate our proposed method, \cref{fig:4_overview} highlights a high-level comparison between ours and previous works, while \cref{fig:5_pipeline} shows our detailed pipeline.

\subsubsection{Stage 1: Recursive Latent Separation}
\label{subsec:method_stage1}

For latent separation, our method utilizes the iterative latent diffusion process, inspired by recent works~\cite{rombach2021latentdiffusion, ke2023marigold}. For the diffusion denoiser, we developed a dual-branch UNet~\cite{unet} to jointly predict the transmission latent and the reflection latent, respectively. As shown in \cref{fig:5_pipeline}, both branches of the UNet are conditioned on the \textit{flash/no-flash latent pair}, which are fixed during the entire diffusion process.

Furthermore, 
{we place cross-attention~\cite{rombach2021latentdiffusion, wang2024stereodiffusion}, between the two branches to iteratively exchange information.}
For both branches, we add zero-initialized cross-attention query and jointly train the two branches to do both self-attention and cross-attention with the opposing branch.
This inter-branch cross-attention
{is important because, given the ground truth latents of the composite images, the predicted transmission latent and the predicted reflection latent can serve as important guidance for the prediction of each other in the next diffusion step.}
\vspace{-2pt}

\begin{figure*}[h!tb]
  \centering
   \includegraphics[width=0.9\textwidth]{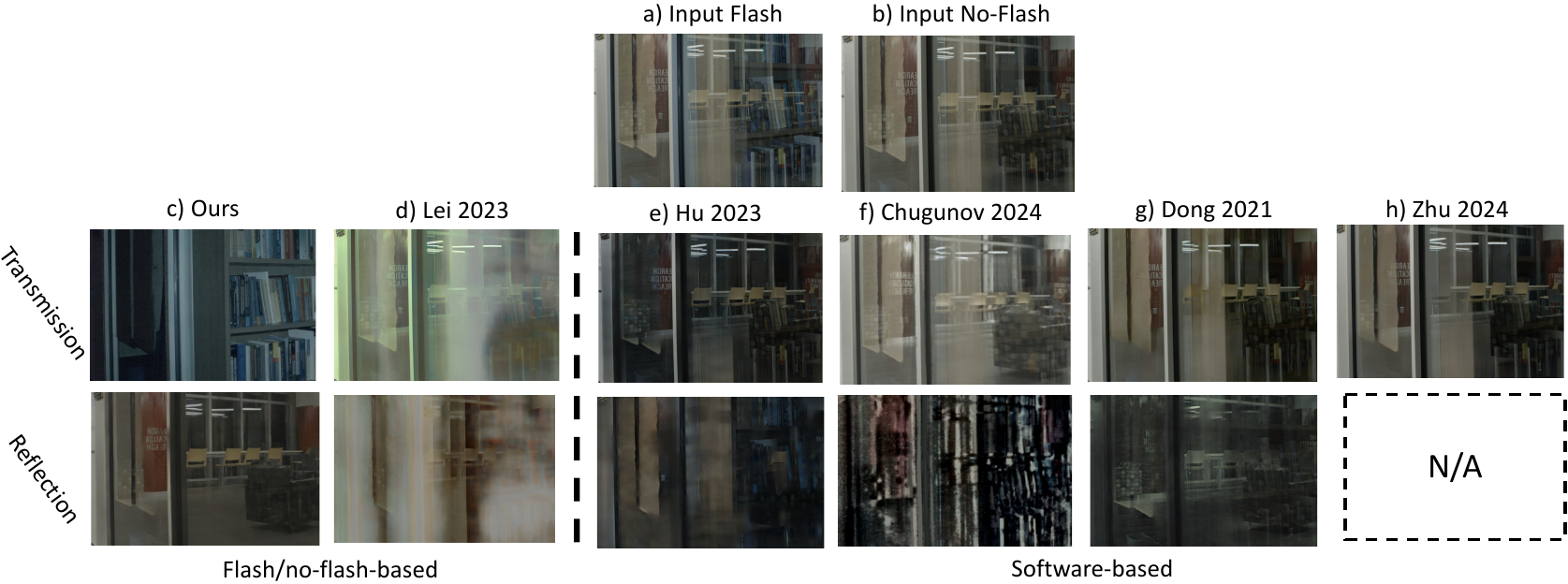}
         \vspace{-5pt}

   \caption{\textbf{Real experiment: The office scene.}  The transmission is a bookshelf inside an office window; the reflection is a study area with chairs and sofas. Our method achieves superior reflection separation than all the baselines (\textcolor{red}{d},\textcolor{red}{e},\textcolor{red}{f},\textcolor{red}{g},\textcolor{red}{h})~\cite{lei2021robust, hu2023single, chugunov2024nsf, dong2021location, zhu2024revisiting}.      
   }
   \label{fig:7_main_office}
         \vspace{-5pt}

\end{figure*}
\begin{figure*}[h!tb]
  \centering
   \includegraphics[width=0.9\textwidth]{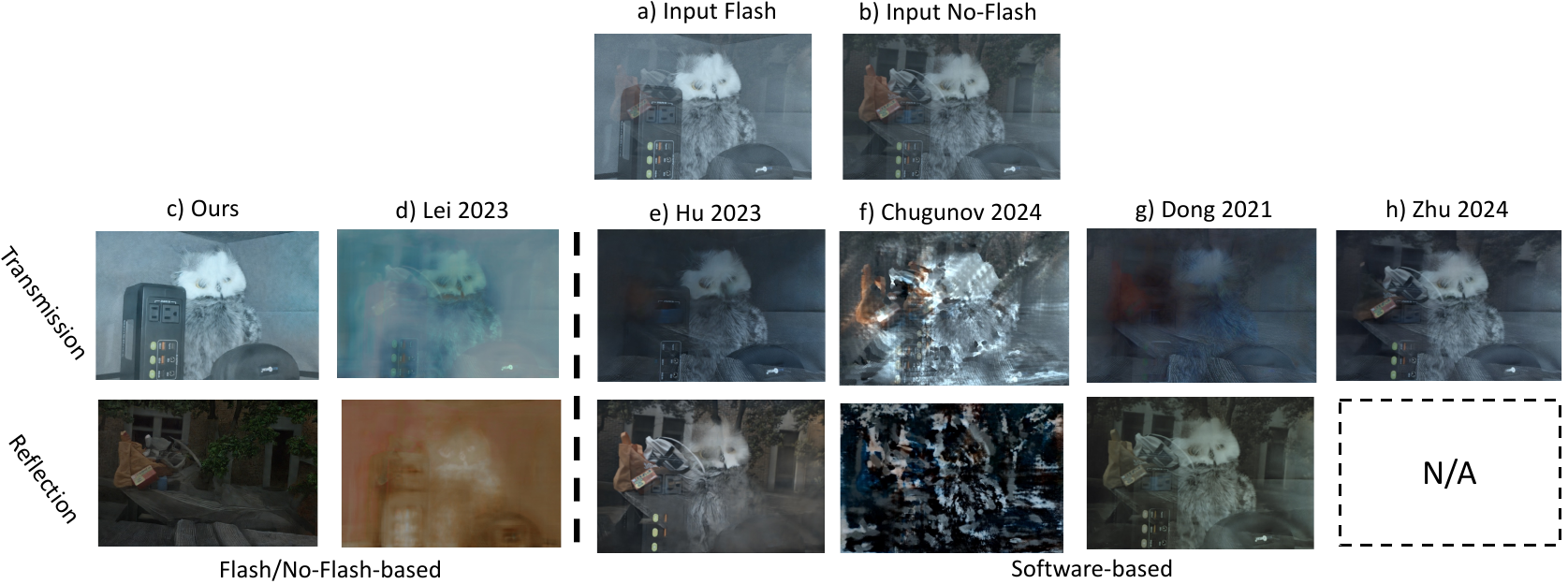}
         \vspace{-5pt}
   \caption{\textbf{Real experiment: the Outdoor Scene.} The transmission is a toy inside a window; the reflection is an outdoor bench.
   Our method achieves superior reflection separation than all the baselines (\textcolor{red}{d},\textcolor{red}{e},\textcolor{red}{f},\textcolor{red}{g},\textcolor{red}{h})~\cite{lei2021robust, hu2023single, chugunov2024nsf, dong2021location, zhu2024revisiting}.  
   }
         \vspace{-5pt}

   \label{fig:7_main_outdoor}
\end{figure*}
\subsubsection{Stage 2: Cross-Latent Decoding}
\label{subsec:method_stage2}

For restoring high-frequency details, the most naive approach is to use a pre-trained VAE decoder~\cite{Kingma2013VAE}; however, we found that the images decoded by it suffer from blurriness, and more importantly, hallucinations (\cref{fig:12_decoder}). 
Given that decoding latent to RGB space is a very under-determined problem,  hallucinations are inevitable unless we supervise the decoding process with other conditions. 
We notice that the original captured image forms a complementary pair with the separated latent from Stage 1: one is unseparated but contains high-frequency features, and one is well-separated but missing high-frequency.
As such, we perform \textit{cross-latent decoding}, as illustrated in \cref{fig:5_pipeline}. 

More specifically, we modify the pre-trained VAE structure to resemble a UNet with zero convolution skip connection but no mid-blocks.
We feed the captured image to the encoder and the separated latent into the decoder.
The zero convolution facilitates stable training and ensures that the trained decoder does not deviate from the original decoder, which contains rich prior information~\cite{zhang2023controlnet}.

With the latent separation and cross-latent decoding stages in place, we now describe the inference procedure for our complete pipeline.
We first encode input flash/no-flash images to obtain input latents, and then concatenate them with a noise latent, before passing through the dual-branch diffusion process to recursively separate apart the latent representations for the transmitted and reflected scenes. 
Once Stage 1 is finished, the separated latents are then passed through our cross-latent decoders with skip connection guidance from unseparated input images, resulting in clear, separated transmission/reflection images.
\vspace{-5pt}


\begin{table}[t]
\vspace{-5pt}
    \centering
    
    \begin{tabular}{lcccc}
    \toprule
         & PSNR $\uparrow$  & SSIM $\uparrow$ &  LPIPS $\downarrow$ & \\
    \midrule
    SDN~\cite{chang2020siamese} & 23.44 & 0.873 & 0.159 &\\
    Lei et al. no align~\cite{lei2023tpami} & 25.41 & 0.917 & 0.112 & \\
      Lei et al.~\cite{lei2023tpami} & 28.60 & 0.956 & 0.071 & \\
      Ours   & \textbf{31.61} & \textbf{0.963} & \textbf{0.048} & \\
      \bottomrule
      
    \end{tabular}
    \caption{\textbf{Quantitative Comparison With Other Flash-based Methods}. This flash/no-flash dataset~\cite{lei2023tpami} contains ground truth images for the transmitted scene, captured by removing the glass. We compare our method against two other flash/no-flash-based methods over the metrics of separated transmission images. Our method achieves significantly better performance.}
    \label{tab:table}
    \vspace{-10pt}
\end{table}
\begin{figure*}[h]
  \centering
   \includegraphics[width=0.8\textwidth]{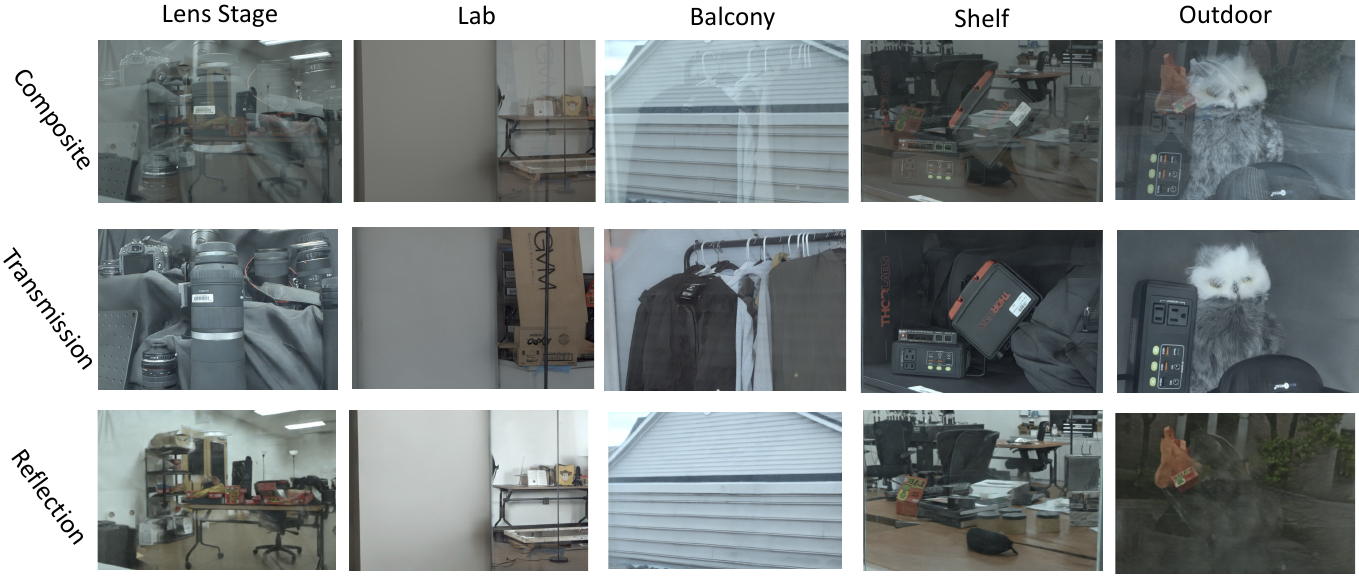}  
   \vspace{-5pt}
   \caption{\textbf{Our Method Works Even Without Access to RAW Images.} Conventional Flash/no-flash methods~\cite{lei2023tpami,xie2024flash-splat} require non-gamma-corrected RAW images as inputs to remove reflections. Likewise, our results shown previously are all using RAW images as input. However, some smartphones, e.g., some models of iPhone, do not give users access to RAW images, which limits the usage of flash/no-flash methods. In this experiment, we train our model to directly take in tonemapped flash/no-flash images as inputs, eliminating the need for RAW images. As shown here, our model still successfully performs reflection separation when applied to real-world tonemapped flash/no-flash image pairs, including the challenging Balcony scene (middle column) with reflections on a double-layer glass door.}
   \label{fig:11_rgb}
      \vspace{-10pt}

\end{figure*}
\section{Experimental Results}
\label{sec:result}
\vspace{-4pt}

\subsection{Experimental Setup}
\vspace{-4pt}
\label{subsec:setup}

Our model is based on the Stable Diffusion architecture ~\cite{rombach2021latentdiffusion}. We used the pre-trained weights of Stable Diffusion 2.1 (SD 2.1) to initialize both of our dual branch and cross-latent decoder.
For dual branch training, we added our inter-branch cross-attention in the midblock attention of the UNet and follow the fine-tuning protocol of Marigold~\cite{ke2023marigold}.
We used the same simulated and real datasets  proposed in ~\citet{lei2023tpami}, which contains sets of flash/no-flash pairs and their corresponding ground truth transmission and reflection. We also evaluated on the data from~\cite{xie2024flash-splat}.
We first trained the dual branch model using a learning rate of $3 \times 10^{-5}$.
We then trained our decoder with output images generated from our Stage 1 model and SD 2.1  decoder, using a learning rate of $10^{-5}$.
Both stages of our model were trained on a NVIDIA A$6000$ GPU, where Stage 1 roughly took two days and Stage 2 took one day.

\textbf{Compared Methods.}
We conducted qualitative and quantitative comparisons using flash/no-flashed-based and pure software-based methods.
We compared with the flash-based method \cite{lei2023tpami}, which consists of an optical flow network to handle misalignment and CNN networks for separation.
We also compared with four recent software-based methods:
\cite{hu2023single}, \cite{dong2021location}, \cite{zhu2024revisiting} are single-image  learning-based methods; while \cite{chugunov2024nsf} is a burst imaging method based on neural rendering.
Our results are presented in Fig~\ref{fig:7_main_lab}, \ref{fig:7_main_poster}, \ref{fig:7_main_office}, \ref{fig:7_main_outdoor}.
Additionally, we compare our results quantitatively with various flash/no-flash-based methods over the real dataset in ~\citet{lei2023tpami}, our method outperforms competing methods in all tested metrics (\cref{tab:table}). 
We finally show additional qualitative comparison with \citet{lei2023tpami} and the ground truth transmission/reflection in \cref{fig:6_sim}.

\begin{figure}[h]
  \centering
   \includegraphics[width=0.9\columnwidth]{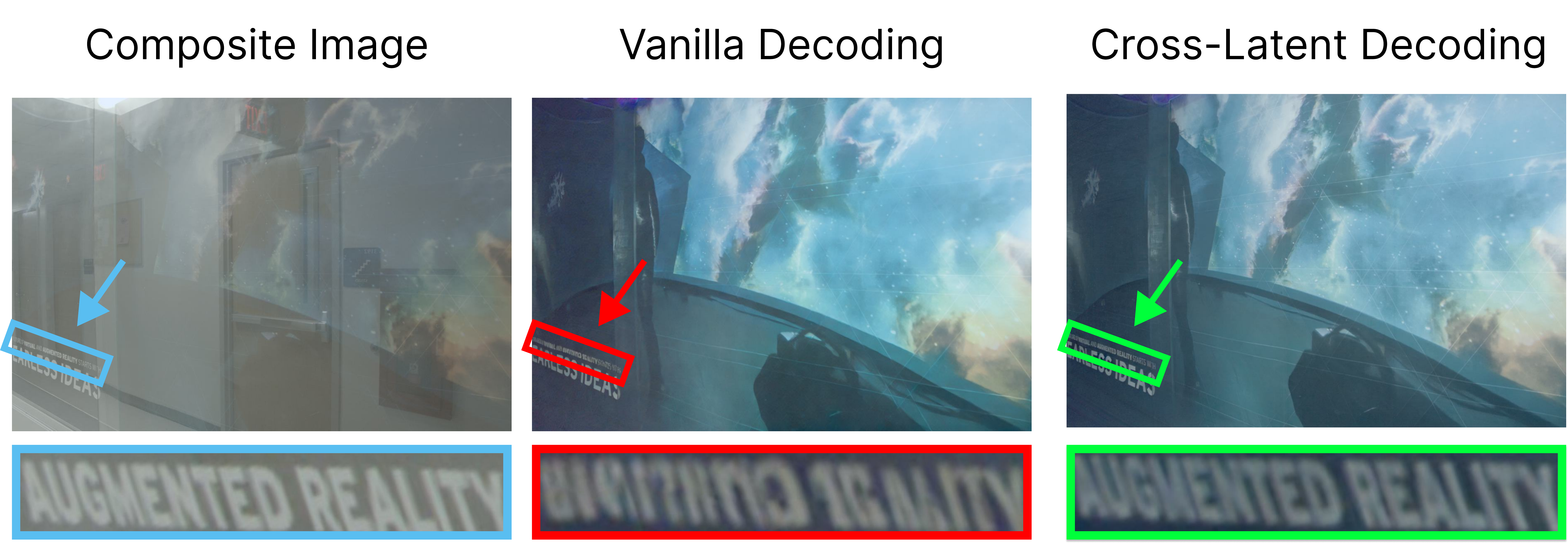}
    \vspace{-5pt}
   \caption{\textbf{Our Cross-Latent Decoder Reduces hallucination}. Compared to the original pre-trained VAE encoder~\cite{Kingma2013VAE}, our cross-latent decoder can leverage the high-frequency signal from the original captured image when it decodes the separated latent. As shown here, our cross-latent decoder's output preserves fine details faithful to the real scene, yielding clearer reconstructions, whereas the vanilla VAE decoder hallucinates the contents.}
   \label{fig:12_decoder}
\end{figure}

\begin{figure}[h]
\vspace{-5pt}
  \centering
   \includegraphics[width=0.9\columnwidth]{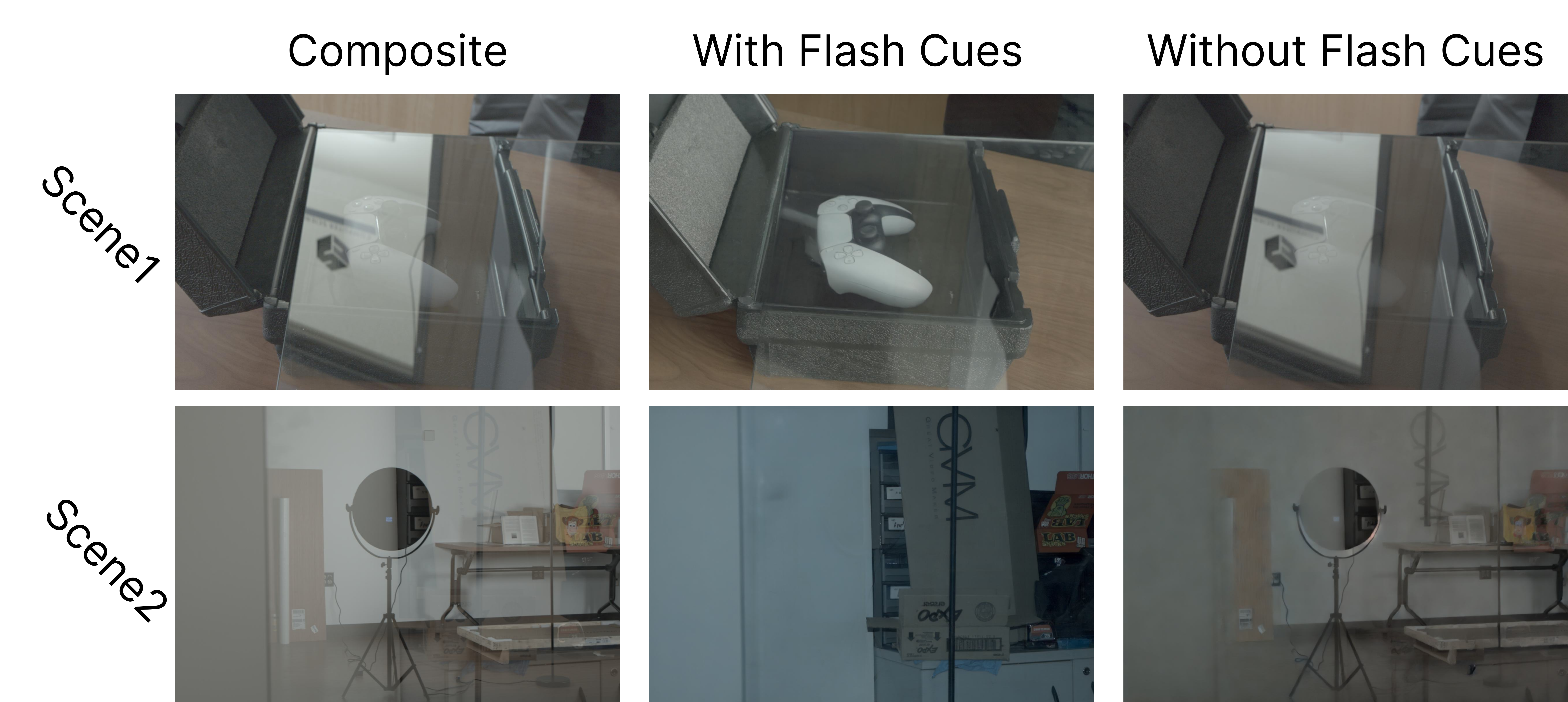}
   \vspace{-5pt}
   \caption{\textbf{Comparing Predicted Transmission of Flash/No-Flash vs. Single Image Models.} We train our model to only take a single composite image as the input. Compared to our flash/no-flash model, the single image model cannot effectively separate reflections. This illustrates the physical cues introduced from the flash/no-flash pair is crucial to our method's success.}
   \label{fig:14_single_input}
\end{figure}

\section{Ablation Studies}
\label{sec:ablation}

\subsection{Reflection Removal with Tonemapped Images}
Flash/no-flash reflection removal methods typically rely on RAW color space image inputs to preserve radiance intensity linearity.
However, access to RAW images is limited on consumer devices (smartphones) without specialized software.
We trained a variant of our model where RAW input is not needed, and instead taking the tonemapped flash/no-flash image as inputs.
We show various real data evaluation results of our model in~\cref{fig:11_rgb}, where \label{subsec:visual_comparison}our model still works with tonemapped image inputs. 
A notable example is the Balcony scene, where the double-layer glass presents a challenging scenario due to multiple reflection paths.
\vspace{-5pt}
\subsection{Cross-Latent Decoder}
\cref{fig:12_decoder} shows that our cross-latent decoder enhances fine details within the separated images.
While our Stage 1 latent separation model effectively isolates the transmission latent feature, the vanilla decoder induces hallucinations and blurriness, e.g., making text illegible as shown in \cref{fig:12_decoder}.
In contrast, our cross-latent decoder is able to extract high frequency details from the composite image based on the predicted latent from Stage 1's separation.

\subsection{Importance of Flash/No-Flash}
\cref{fig:14_single_input} shows that the flash cues are crucial for latent-space separation: we trained a variant of our model that takes in a single image as the input (instead of the flash/no-flash pair), and it failed to remove the reflections.
\vspace{-5pt}

\section{Conclusions}
\label{sec:conclusions}
In conclusion, our Flash-Split method provides a robust solution for reflection separation in transparent surfaces, overcoming the need for precise flash/no-flash alignment. By performing reflection separation in the latent space, we effectively circumvent the flash/no-flash misalignment issue. We also employ a cross-latent decoding module to restore detailed and faithful features of the separated scenes from their latents. Evaluations on both simulated and challenging real-world data confirm our effectiveness, marking a substantial improvement in practical reflection separation.

{
    \small
    \bibliographystyle{ieeenat_fullname}
    \bibliography{main}
}

\clearpage
\setcounter{page}{1}
\setcounter{figure}{13}
\setcounter{equation}{3}
\maketitlesupplementary

In this supplementary material, we evaluate an additional flash/no-flash baseline~\cite{chang2020siamese} on real scenes (\cref{sec:sup_sdn}), demonstrate the respective roles of the two stages of our method (latent separation and cross-latent decoding) (\cref{sec:sup_stage_1_vs_2}), analyze our method's robustness to misalignment (\cref{sec:sup_misalignment}), report baseline's performance on flash images (\cref{sec:sup_flash}), and provide more training and inference details (\cref{sec:sup_training}). 
\section{Additional Flash/No-Flash-Based Baseline}
\label{sec:sup_sdn}
In our main paper we compared our results with a flash/no-flash baseline \citet{lei2023tpami}, this is the most recent method on flash/no-flash based reflection separation method. We take \citet{lei2023tpami}'s official code implementation from GitHub for their method and use their pretrained network checkpoints.
However, as shown in \cref{fig:7_main_lab}, \ref{fig:7_main_poster}, \ref{fig:7_main_office}, \ref{fig:7_main_outdoor} of the main paper, the reflection separation performance of \citet{lei2023tpami} are not satisfactory.

We additionally add another flash/no-flash baseline, \citet{chang2020siamese},  which proposes a siamese dense network (SDN) for reflection removal with flash and no-flash image pairs. We also use their official implementation plus their pretrained checkpoints.
We evaluate this method using the same scenes shown in the main paper.
The results are shown in \cref{fig:X_main_lab}, \ref{fig:X_main_poster}, \ref{fig:X_main_office}, \ref{fig:X_main_outdoor}. 
These four scenes correspond to \cref{fig:7_main_lab}, \ref{fig:7_main_poster}, \ref{fig:7_main_office}, \ref{fig:7_main_outdoor} of the main paper.
While \citet{chang2020siamese} outperforms \citet{lei2023tpami} on real data, it still falls short of fully separating the transmission component from the input flash and no-flash images.
Our method still achieves much better reflection separation performance.

\begin{figure}[t]
  \centering
   \includegraphics[width=\columnwidth]{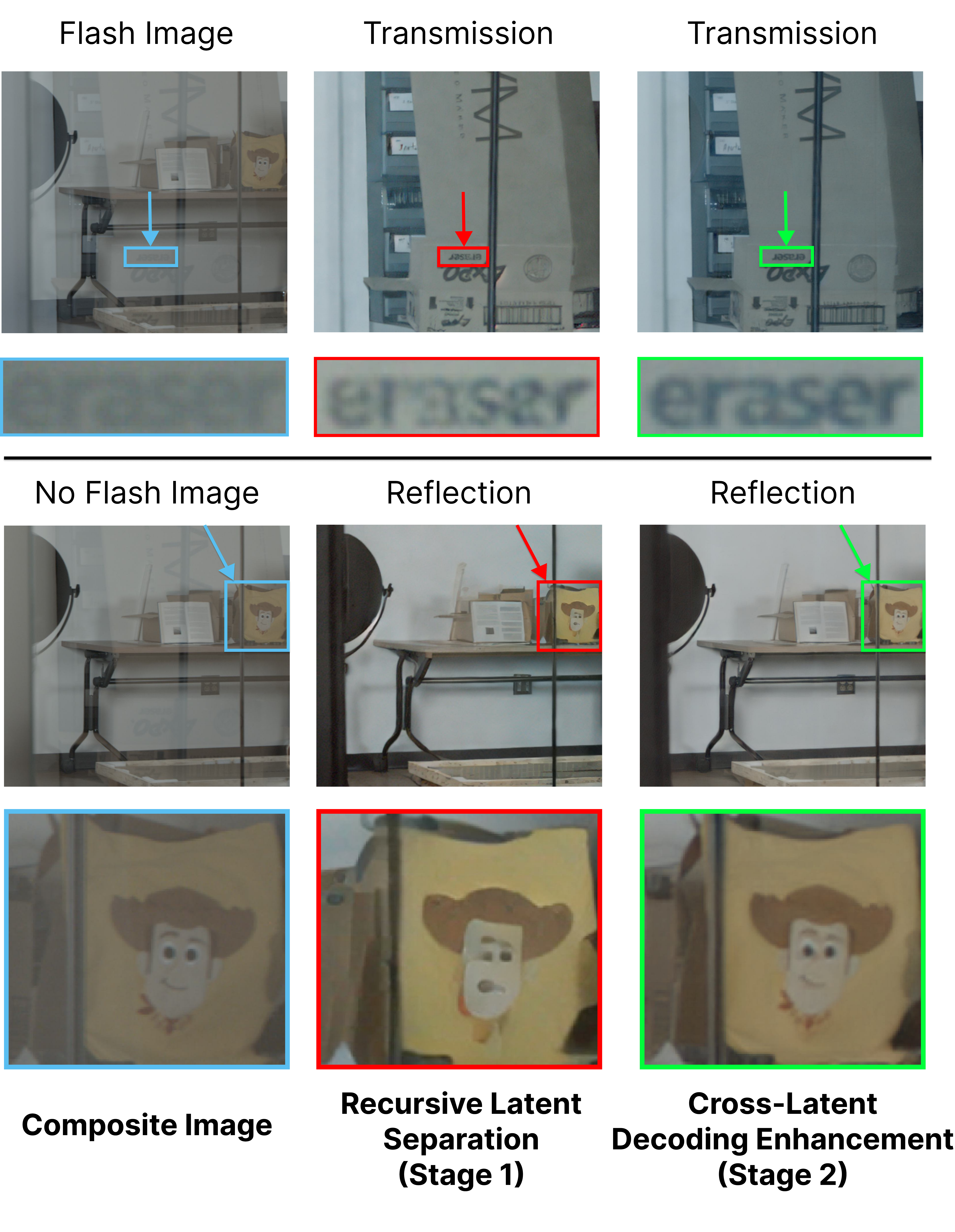}
   \vspace{-10 pt}

   \caption{\textbf{Stage 1 for Separation; Stage 2 for Enhancement}. We visualize the intermediate results from our Recursive Latent Separation in Stage I (middle column) and the final results from our Cross-latent Decoding in Stage 2 (right column).
   {Stage 1 of our method performs good separation, and Stage 2 enhances the details while avoiding hallucinations.} Note that in our method, Stage 1 only outputs the separated transmission/reflection latents, but in this figure, for the purpose of visualization, we decode the separated latents using a vanilla decoder from~\cite{rombach2021latentdiffusion}. Additionally, note that the zoom-in texts  (``eraser") shown in the top half of this figure have been flipped vertically for better readability.
   }
   \label{fig:X_stage_2_lab}
   \vspace{-10pt}
\end{figure}
\begin{figure}[t]
  \centering
   \includegraphics[width=\columnwidth]{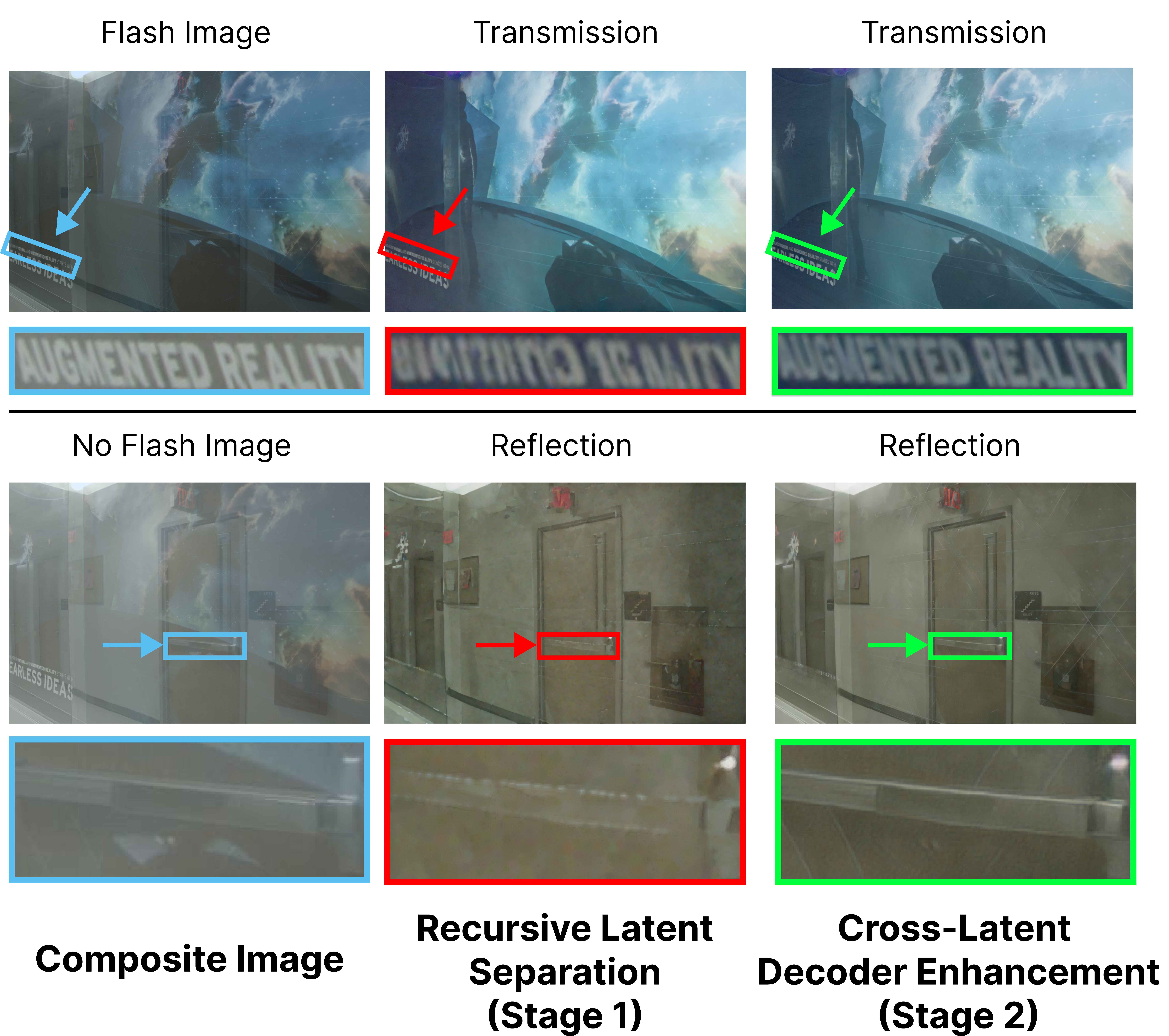}
   \vspace{-10pt}
   \caption{\textbf{Stage 1 for Separation; Stage 2 for Enhancement}. Same experiment as \cref{fig:X_stage_2_lab}, but on a new scene. 
   Stage 1 of our method perform good separation, and Stage 2 enhances the details while avoiding hallucinations. 
   Note that the two small white triangles in the zoomed-in regions of the captured composite no-flash image (lower left corner) are from the transmitted scene, which aligns with our model's prediction.
   }
   \label{fig:X_stage_2_poster}
   \vspace{-10pt}
\end{figure}
\section{Respective Roles of Our 2-Stage Separation}
\label{sec:sup_stage_1_vs_2}

As mentioned in our main paper, we decouple the reflection separation problem into two consecutive stages: (1) recursive latent separation and (2) cross-latent decoding. 
More specifically, in Stage 1, we recursively separate the reflection and transmission within the latent space; in Stage 2, we restore fine image details to the separated latents while keeping the reconstruction faithful to the original scene, by using separated latent from Stage 1 as guidance to extract the sharp image features from the unseparated input image.

The respective effects of the two stages are shown in \cref{fig:X_stage_2_lab} and \ref{fig:X_stage_2_poster}. 
To visualize the intermediate results after Stage 1 (recursive latent separation), we decode the separated transmission/reflection latents using a vanilla decoder~\cite{rombach2021latentdiffusion}. We can clearly see that the recursive latent separation in Stage 1 already performs a good separation of reflection and transmission. However, these intermediate results still suffer from hallucinations and blurriness, due to the under-determinedness of the decoding process. In Stage 2, our cross-latent decoding significantly improves the sharpness and faithfulness of the reconstructed images by leveraging the high-frequency details contained in the original input images. 

In summary, Stage 1 separates the transmission and reflection, while Stage 2 enhances the details.

\begin{figure*}[t]
  \centering
   \includegraphics[width=\textwidth]{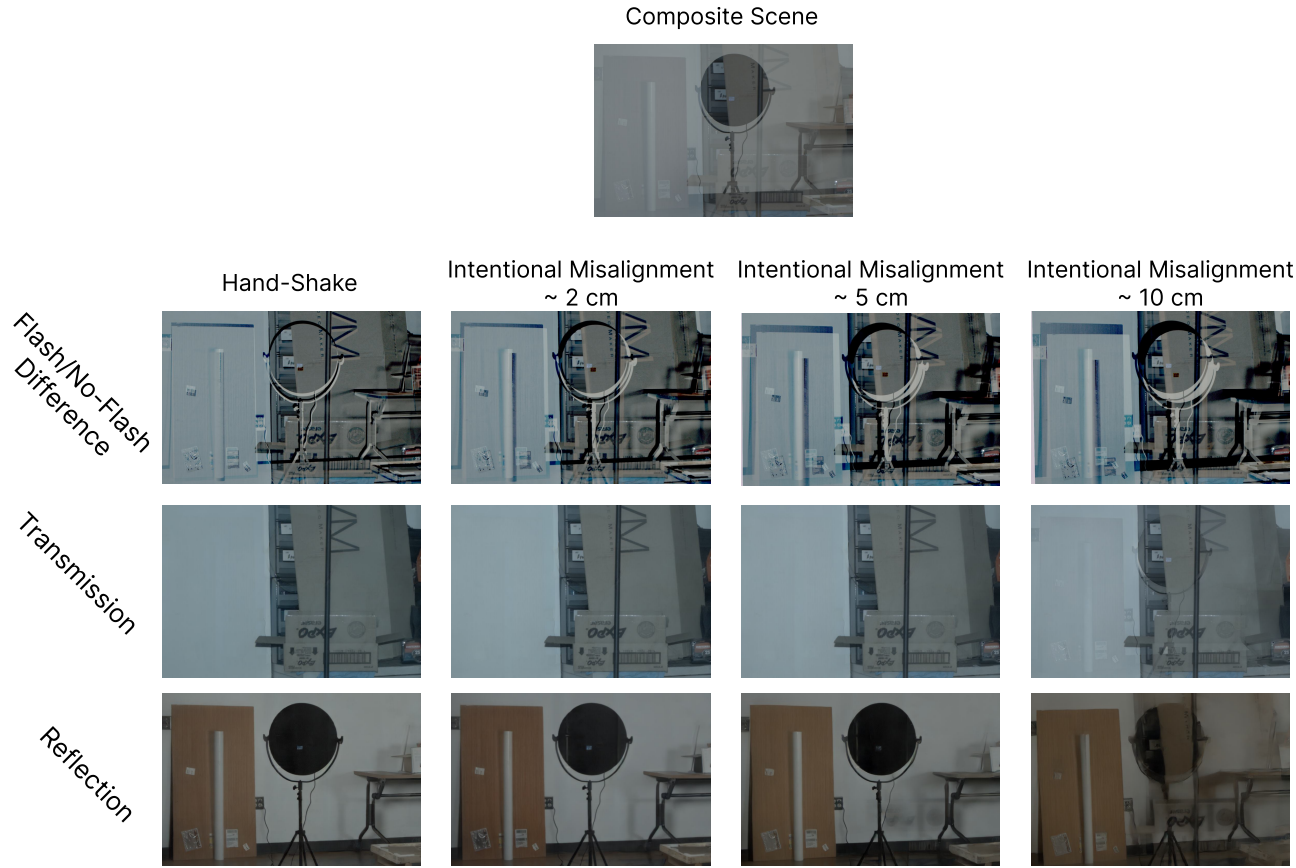}
   \caption{\textbf{Our Method's Robustness to Different Degrees of Misalignment}. While our model effectively handles misaligned flash and no-flash images due to handshake, we intentionally further increase the flash/no-flash misalignment to find out when will our model fail. Note that we assume the scene to be static and the misalignment comes from camera motion.
   \textcolor{black}{The results show our method's robustness against estimated misalignment of 2 and 5 centimeters, respectively.
   However, when the misalignment exceeds 10 centimeters, our method's performance deteriorates. } 
   This shows that our method performs robustly against small to moderate camera motion (e.g., hand shake) while baselines completely fail; however, in the case of very severe camera motion, (e.g., if the user is running or biking while capturing the flash/no-flash pair), our method might fail.
   }
   \label{fig:X_misalignment}
\end{figure*}

\vspace{-5pt}
\section{Robustness Against Misalignment}
\label{sec:sup_misalignment}

To better understand our model's robustness to more severe misalignment, we intentionally increase the amount of misalignment between the captured flash and no-flash images, to a degree where our method fails. 
\cref{fig:X_misalignment} shows that our method performs robustly against small to moderate camera motion (e.g., hand shake)
; however, in the case of extreme camera motion, (e.g., if the user is running or biking while capturing the flash/no-flash pair), our method might fail.

\section{Software-based Methods Using Flash Image}
\label{sec:sup_flash}
The goal of this section is to show that our method performs better not because we use a camera flash, but rather because we use the cues from the flash/no-flash difference.

In our main paper, we visually compared our method with various software-based reflection removal methods. 
In those comparisons, we fed the {no-flash} images as inputs to the software-based methods.
The rationale behind this choice is that these methods were trained on no-flash images, making the no-flash inputs in our real image evaluation more representative of their training distribution. 
Consequently, we believe that this approach provides a fair baseline for comparison.

\begin{figure*}[h!tb]
  \centering
   \includegraphics[width=0.9\textwidth]{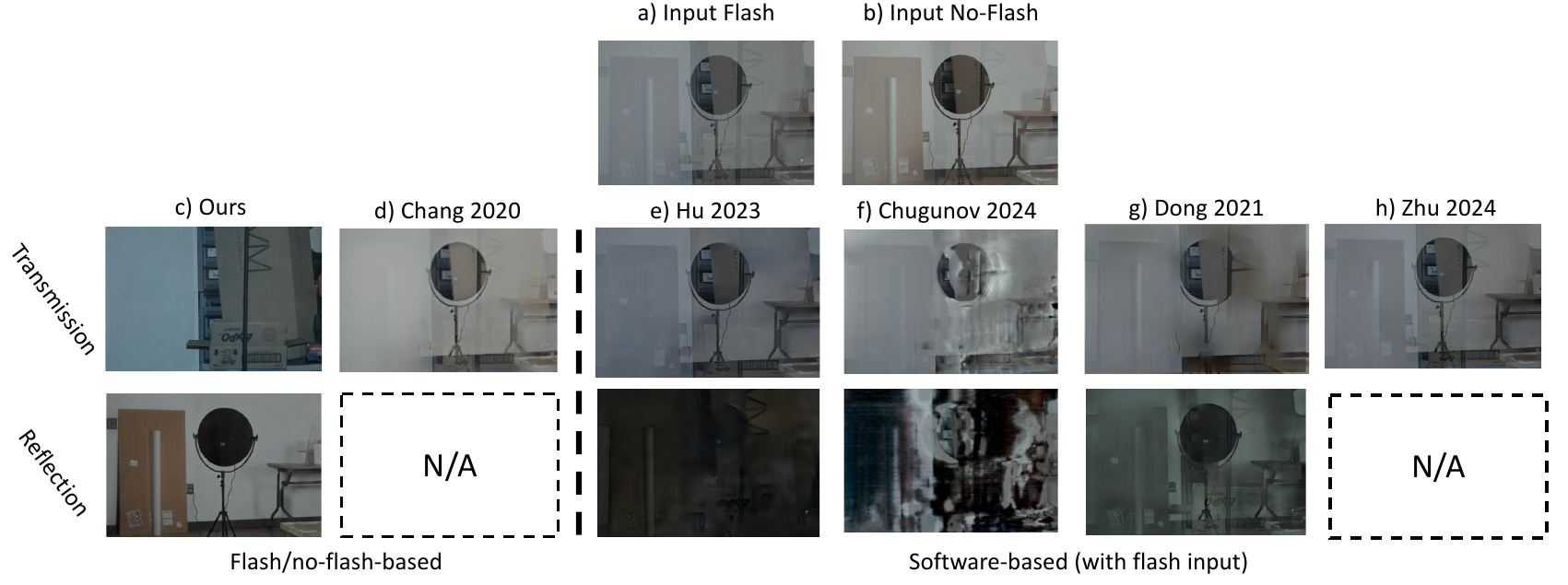}
      \vspace{-5pt}

   \caption{\textbf{Real Experiment: The Lab Scene.} 
      We compare with an additional flash/no-flash-based baseline \citet{chang2020siamese}.
   \citet{chang2020siamese} can only predict the transmission, not the reflection, thus the ``N/A''.
    Although \citet{chang2020siamese}  achieves better results than \citet{lei2023tpami} on the real data, it still cannot completely separate the transmission component from the input flash/no-flash images.
   The software-based results shown in the real experiment are obtained using the no-flash image as the input.
       This figure provides additional results to \cref{fig:7_main_lab} of the main paper.
    }
   \vspace{-5pt}
   \label{fig:X_main_lab}
\end{figure*}

\begin{figure*}[ht]
  \centering
   \includegraphics[width=0.9\textwidth]{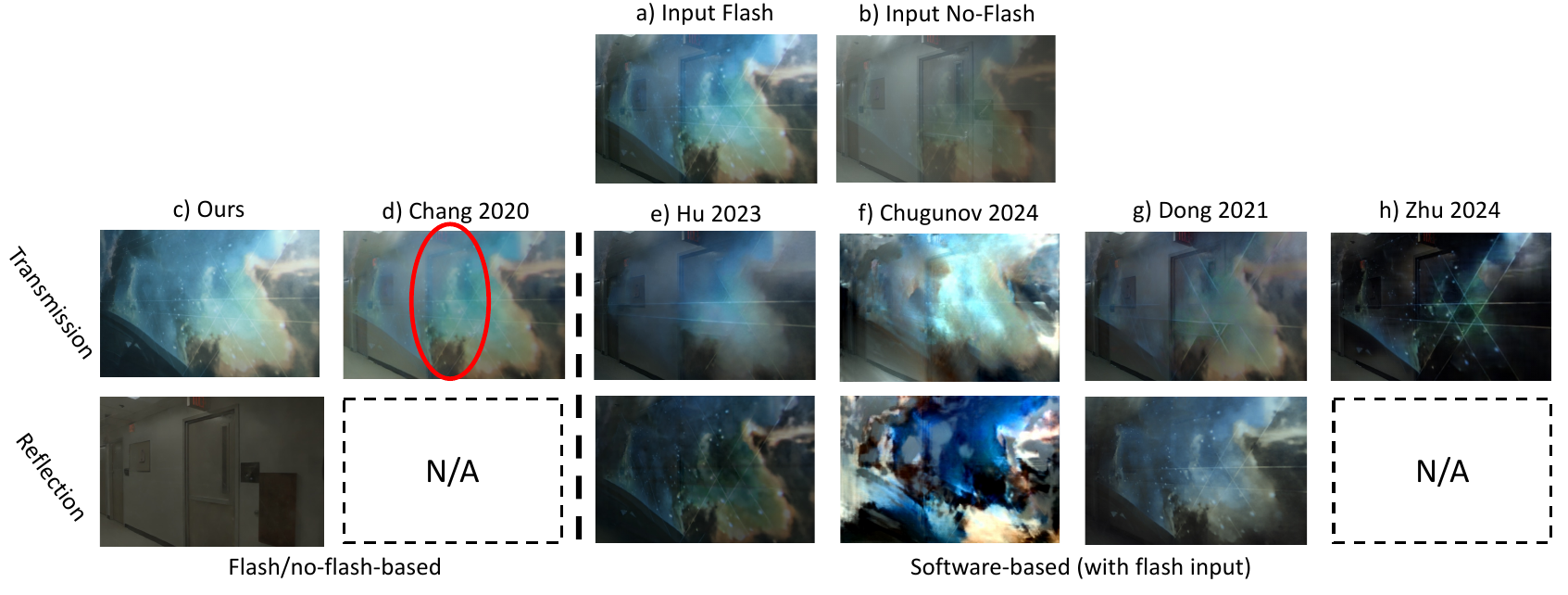}
   \vspace{-5pt}
   \caption{\textbf{Real Experiment: The Poster Scene.}
   We compare with an additional flash/no-flash-based baseline \citet{chang2020siamese}.
   \citet{chang2020siamese} can only predict the transmission, not the reflection, thus the ``N/A''.
    Although \citet{chang2020siamese}  achieves better results than \citet{lei2023tpami} on the real data, it still cannot completely separate the transmission component from the input flash/no-flash images.
    We circle the areas where \citet{chang2020siamese} did not correctly separate the door in the reflection. 
   The software-based results shown in the real experiment are obtained using the no-flash image as the input.
    This figure provides additional results to \cref{fig:7_main_poster} of the main paper.
   }
   \label{fig:X_main_poster}
   \vspace{-10pt}
\end{figure*}

\begin{figure*}[h!tb]
  \centering
   \includegraphics[width=0.9\textwidth]{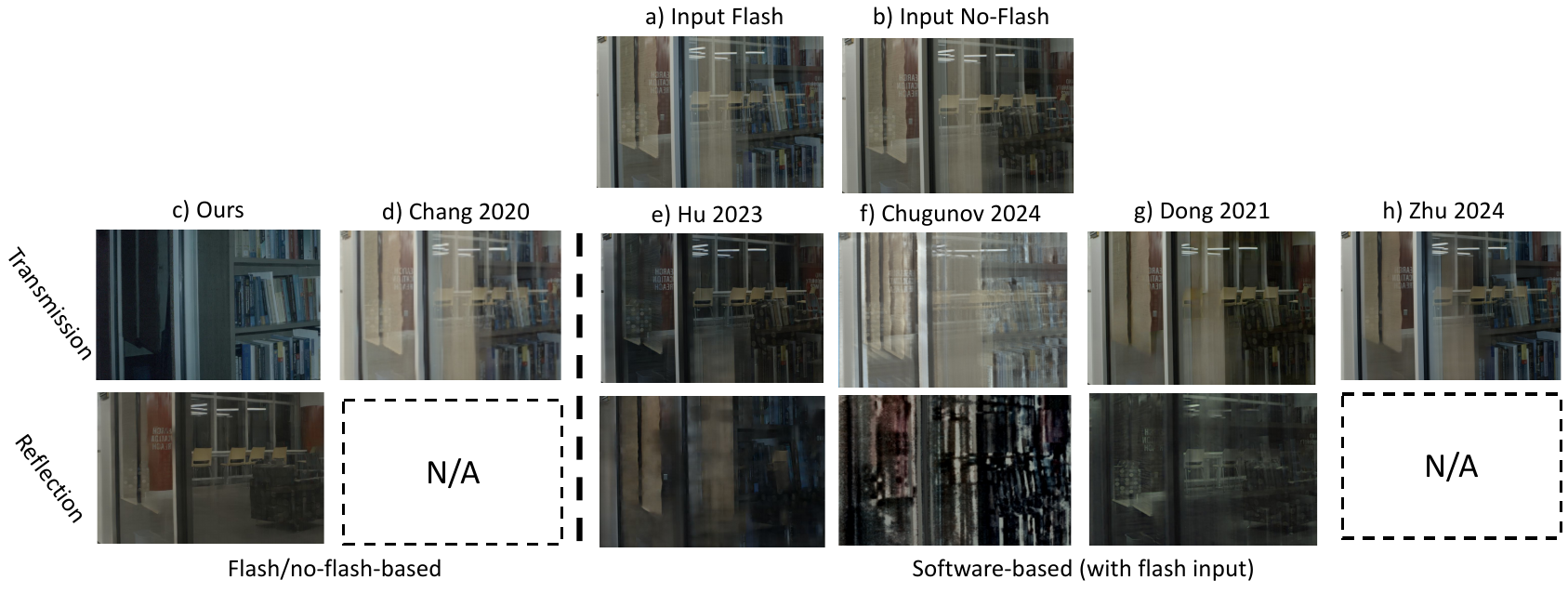}
         \vspace{-5pt}
   \caption{\textbf{Real experiment: the Office Scene.}
      We compare with an additional flash/no-flash-based baseline \citet{chang2020siamese}.
   \citet{chang2020siamese} can only predict the transmission, not the reflection, thus the ``N/A''.
    Although \citet{chang2020siamese}  achieves better results than \citet{lei2023tpami} on the real data, it still cannot completely separate the transmission component from the input flash/no-flash images.
   The software-based results shown in the real experiment are obtained using the no-flash image as the input.
    This figure provides additional results to \cref{fig:7_main_office} of the main paper.
   }
         \vspace{-5pt}

   \label{fig:X_main_office}
\end{figure*}

\begin{figure*}[ht]
  \centering
   \includegraphics[width=0.9\textwidth]{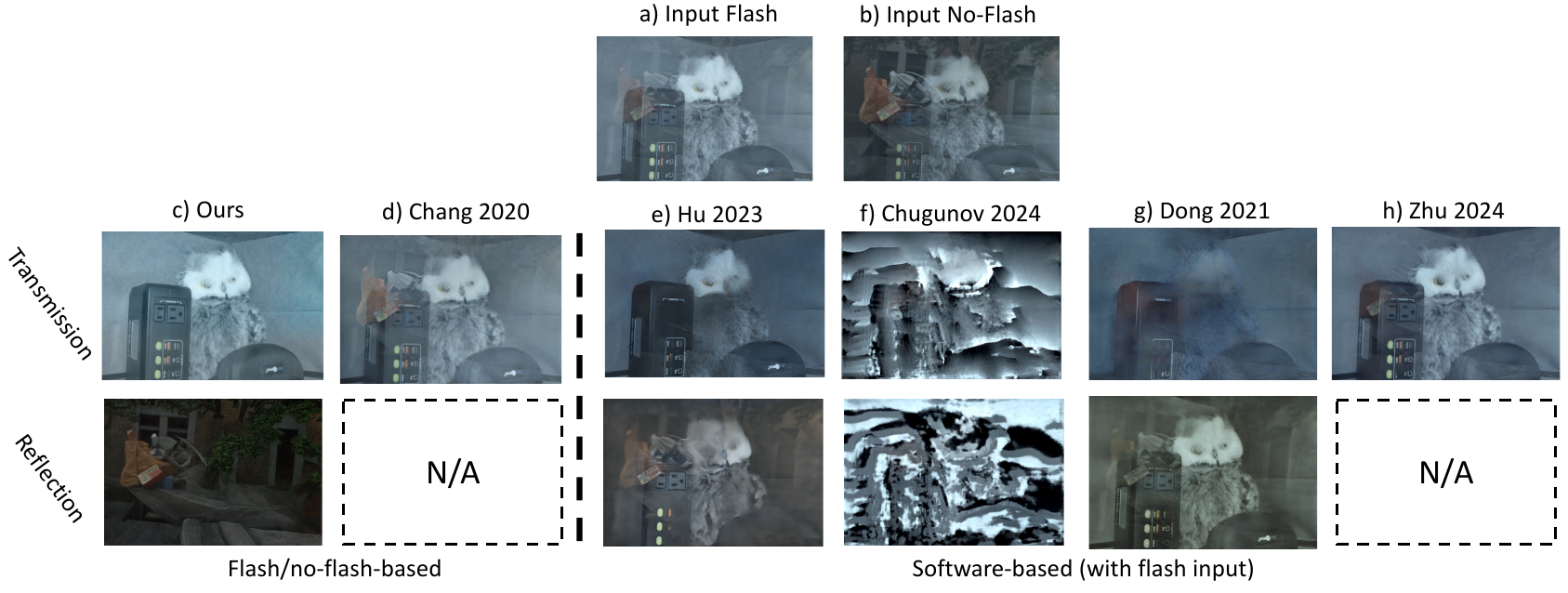}
         \vspace{-5pt}
   \caption{\textbf{Real experiment: the Outdoor Scene.}
   \citet{chang2020siamese} can only predict the transmission, not the reflection, thus the ``N/A''.
    Although \citet{chang2020siamese}  achieves better results than \citet{lei2023tpami} on the real data, it still cannot completely separate the transmission component from the input flash/no-flash images.
   The software-based results shown in the real experiment are obtained using the no-flash image as the input.
    This figure provides additional results to \cref{fig:7_main_outdoor} of the main paper.
   }
         \vspace{-5pt}

   \label{fig:X_main_outdoor}
\end{figure*}

However, one could argue that the flash images, which exhibit a stronger transmission component, might provide an advantage for software-based methods to better separate out the transmissions. 
To address this potential concern, we additionally run software-only baselines on the same scenes shown in the main paper, but using the \textit{flash images} as inputs. The results are shown in \cref{fig:X_main_lab}, \ref{fig:X_main_poster}, \ref{fig:X_main_office}, \ref{fig:X_main_outdoor}. 
These four scenes correspond to \cref{fig:7_main_lab}, \ref{fig:7_main_poster}, \ref{fig:7_main_office}, \ref{fig:7_main_outdoor} of the main paper.

In this case, our method still achieves much better reflection separation performance compared to software-based methods, which implies that, compared to the software-based methods, our method's superiority does not come from flash, but rather, comes from the flash/no-flash cues. 


\vspace{-10pt}
\section{Additional Training and Inference Details}
\label{sec:sup_training}
In this section, we provide additional details on the training and inference procedures for Stage 1 latent separation and Stage 2 cross-latent decoding. 
At a high level, our proposed pipeline is introduced in \cref{fig:5_pipeline} and \cref{subsec:method_pipeline} of the main paper.

\subsection{Training}

\noindent\textbf{Stage 1 Latent Separation.} 
During Stage 1 latent separation training, we convert the flash and no-flash images to the latent space using the vanilla encoder from~\cite{rombach2021latentdiffusion}, and concatenate them in the latent channel dimension to form an input latent image $z$.
We then take the target ground truth transmission/reflection images and encode them into ground truth image latents $s_{0}$. 
Now, we sample a noise image latent $\epsilon$ with the same dimension as the ground truth image latent.
We then add the noise image to the ground truth image latent using a random noise level $t$:
\begin{equation}
    s_t =\sqrt{\alpha_t} s_0 + (\sqrt{1 - \alpha_t}) \epsilon
\end{equation}
Here $\{\alpha_t\}, t \in \{1,..., T\}$ is the noise schedule specific to the diffusion model. 
We use the default DDPM~\cite{ho2020denoising_ddpm} scheduler of the Stable Diffusion 2.1 model~\cite{rombach2021latentdiffusion} with $T =1000$ steps for training.
We also use the annealed multi-resolution noise~\cite{ke2023marigold} instead of standard Gaussian noise~\cite{ho2020denoising_ddpm}. 

Our UNet then takes the input latents from the flash/no-flash input images $(z)$ and noised ground truth latents $(s_t)$ from the ground truth transmission/reflection images
and predicts a noise $\hat{\epsilon}$.
Our training objective is to minimize the L2 loss between the injected noise $\epsilon$ and the noise predicted by the UNet $\hat{\epsilon}$. Note that the ground truth transmission/reflection images are only used for training, and not used for inference (see \cref{sec:X_inference} for details).

We use the exact simulated and real datasets as proposed in ~\citet{lei2023tpami}, which contains sets of flash/no-flash pairs and corresponding ground truth transmission and reflection images. The input images are randomly cropped to $384 \times 384$ sized patches for training. To simulate misaligned flash/no-flash image pairs, we follow~\citet{lei2023tpami} and keep the no-flash images intact and do a monocular-depth-guided image misalignment to generate a misaligned flash image. \\

\noindent\textbf{Stage 2 Cross-Latent Decoder.} 
Our cross-latent decoder is trained to learn a mapping from the latents separated by our Stage 1 (recursive latent separation) to the ground truth transmission/reflection images, using unseparated input images as guidance.

The architecture of our cross-latent decoder is modified from the pre-trained VAE component in~\cite{rombach2021latentdiffusion} by adding skip connections with zero convolutions.
We trained separate cross-latent decoders for reflection and transmission.
For transmission, we use the input flash image as the composite image, since the flash image contains a higher proportion of transmission compared to the no-flash image.
Conversely, for reflection, we use the input no-flash image as the composite image, since the no-flash image contains a higher proportion of reflection compared to the flash image.
Our cross-latent decoder takes in both the unseparated input image and the separated latent from Stage 1 as inputs, and outputs a separated RGB image.
We train the model by minimizing the difference between the decoded and the ground truth transmission/reflection image.
We use an equally weighted sum of L1, SSIM~\cite{Baker:DSSIM}, and LPIPS~\cite{zhang2018unreasonable} losses to supervise the training.
We take the separated transmission/reflection latents from Stage 1 and group them with the ground truth transmission/reflection, as well as the input flash/no-flash images to form our Stage 2 training data.
Specifically, we take the misaligned training images crops with size $384 \times 384$ from \citet{lei2023tpami} as the input and run inference on our trained Stage 1 model for 20 DDIM~\cite{song2020denoising_ddim} denoise iterations. 
See \cref{sec:X_inference} on Stage 1 inference
for more details.

\subsection{Inference}
\label{sec:X_inference}
{After training, our diffusion model can be used to recover transmission/reflection images from any flash no-flash pair.}
We convert the flash and no-flash images to the latent space using the vanilla encoder from~\cite{rombach2021latentdiffusion}, and concatenate them in the latent channel dimension to obtain the input latent image  $z$.
Our output prediction latent for transmission/reflection $s$ is initialized from random Gaussian noise.
We iteratively denoise the separated reflection/transmission images using our trained dual-branch UNet under the guidance of input flash/no-flash images.
At each denoising iteration, we concatenate the input and output prediction latent images and feed them to the UNet. We then update the prediction latent based on the predicted noise of our UNet and the current time step.
\begin{equation}
    s_{t - 1} = DDIM(s_t, \hat{\epsilon}_t, t)
\end{equation}
Here $t$ is the denoising timestep for the current iteration.
This denoising timestep corresponds to the amount of noise contained in the output latent and decreases with every subsequent denoising iteration. 
$s_t$ is the output separated transmission/reflection image at timestep $t$, $\hat{\epsilon_t}$ is the noise predicted by the UNet at timestep $t$, and $s_{t - 1}$ is the output separated transmission/reflection latent image at the next timestep $t - 1$ ready for the next iteration. 
We use the DDIM~\cite{song2020denoising_ddim} scheduler for inference, which uses skipping step updates to enable fewer denoising iterations and faster inference. 
We use $50$ denoising iterations for inference.

{Inference continues to Stage 2 where we take the separated transmission/reflection latent outputs from Stage 1 and feed them to the decoder of their respective cross-latent decoders. }
Finally, the Stage 2 cross-latent decoders output the refined transmission/reflection RGB images.

\end{document}